\documentclass[letterpaper, 10 pt, conference]{ieeeconf}  

\IEEEoverridecommandlockouts                              

\title{\LARGE \bf
SCU-Hand with Integrated Single-Sheet Valve: 
\\
A Funnel-Shaped Robotic Hand for Milligram-Scale Powder Handling
}

 \author{Tomoya Takahashi$^{1}$, Yusaku Nakajima$^{2}$, Cristian C. Beltran-Hernandez$^{1}$, Yuki Kuroda$^{1}$, \\ Kanta Ono$^{2}$, Kazutoshi Tanaka$^{1}$, Masashi Hamaya$^{1}$, and Yoshitaka Ushiku$^{1}$
 \thanks{*This work was supported by the JST-Mirai Program Grant Number JPMJMI21G2, Japan.}
 \thanks{$^{1}$OMRON SINIC X Corporation, Tokyo, Japan. $^{2}$The University of Osaka, Osaka, Japan.
         {\tt\small tomoya.takahashi@sinicx.com}}%
 }

\usepackage[dvipdfmx]{graphicx}
 \usepackage{multirow}
 \usepackage[table,xcdraw]{xcolor}
 \usepackage{textcomp, gensymb}
\usepackage{amsmath}

\usepackage{color}
\usepackage{colortbl}

\usepackage[absolute,overlay]{textpos}
\usepackage{algorithm}
\usepackage{algorithmicx}
\usepackage{algpseudocode}
\usepackage{amsmath}
\usepackage{float}
\usepackage{booktabs} 
\begin{document}

\maketitle
\thispagestyle{empty}
\pagestyle{empty}

\begin{textblock*}{18cm}(1.5cm,0.5cm) 
    \tiny 2026 IEEE International Conference on Robotics and Automation (ICRA2026). Preprint. Accepted January 2026. \textcopyright 2026 IEEE.  Personal use of this material is permitted.  Permission from IEEE must be obtained for all other uses, in any current or future media, including reprinting/republishing this material for advertising or promotional purposes, creating new collective works, for resale or redistribution to servers or lists, or reuse of any copyrighted component of this work in other works.
\end{textblock*}

\begin{abstract}
Laboratory Automation (LA) has the potential to accelerate solid-state materials discovery by enabling continuous robotic operation without human intervention. 
While robotic systems have been developed for tasks such as powder grinding and X-ray diffraction (XRD) analysis, fully automating powder handling at the milligram scale remains a significant challenge due to the complex flow dynamics of powders and the diversity of laboratory tasks. To address this challenge, this study proposes the SCU-Hand-SV (Soft Conical Universal Robotic Hand with Single-sheet Valve), which preserves the softness and conical sheet designs in prior work while incorporating a controllable valve at the cone apex to enable precise, incremental dispensing of milligram-scale powder quantities. The SCU-Hand-SV is integrated with an external balance through a feedback control system based on a model of powder flow and online parameter identification. 
Experimental evaluations with glass beads, monosodium glutamate, and titanium dioxide demonstrated that 80\% of the trials achieved an error within $\pm$2\,mg, and the maximum error observed was approximately 20\,mg across a target range of 20\,mg to 3\,g. In addition, by incorporating flow prediction models commonly used for hoppers and performing online parameter identification, the system is able to adapt to variations in powder dynamics. Compared to direct PID control, the proposed model-based control significantly improved both accuracy and convergence speed. These results highlight the potential of the proposed system to enable efficient and flexible powder weighing, with scalability toward larger quantities and applicability to a broad range of laboratory automation tasks.

\end{abstract}


\section{Introduction}
Laboratory Automation (LA) has the potential to accelerate solid-state materials discovery significantly by enabling robots to operate continuously, 24 hours a day, without human intervention~\cite{duros2017human}. While many products and systems have been developed, the complete automation of small-scale, diverse solid-state materials discoveries remains a significant challenge~\cite{christensen2021automation}, even though specific processes, such as weighing~\cite{kadokawa2023learning}, solid-state synthesis workflows including sintering~\cite{omidvar2024accelerated}, and loading samples into X-ray diffraction (XRD) analysis~\cite{Yotsumoto2024} instruments, have already been automated with structurally similar robotic systems.
In particular, transferring powders and containers between instruments cannot be fully handled by dedicated machines alone, and the necessity of human involvement often becomes a bottleneck for continuous operation. 
For such small-scale LA tasks, general robotic systems such as multi-degree-of-freedom manipulators are expected to play an important role~\cite{tom2024self}. 
For this reason, this study focuses on developing robotic methods for handling powder, a critical but unsolved task of small-scale LA.

However, scooping powders from a container and transferring a defined amount into another container involves multiple challenges for robotic systems. In scooping, it is often necessary to collect nearly all of the powder inside the container in order to secure sufficient material. Humans achieve this easily by sliding a rigid spatula along the container wall, but replicating such a motion requires robots to employ force control, visual sensing, or reinforcement learning~\cite{pizzuto2022accelerating}.  
In the dispensing phase, realizing milligram-scale precision demanded in laboratory practice requires robots to perform milligram-scale dumping of powders with extremely high joint accuracy, as well as to employ control strategies capable of adapting to the complex dynamics of powder flow~\cite{tuomainen2022manipulation}. Furthermore, the system must handle a wide range of target quantities and container types. Powders ground in a mortar may need to be transferred into XRD holders, pellet dies, or crucibles for sintering, each requiring amounts that range from several tens of milligrams to several grams. Humans adapt to these diverse requirements by freely using different spatulas or tools, but for robots, tool switching and decision making remain major issues.  


   \begin{figure}[t]
      \centering
     \includegraphics[width=0.9\linewidth]{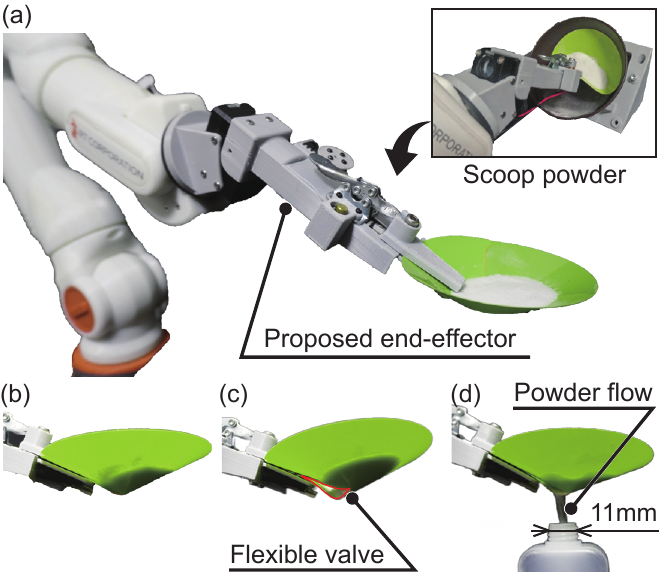}
      \caption{SCU-Hand-SV: (a)~Scooping granular media, (b)~Bottom view of flexible valve, (c)~open valve, (d)~Dispensing into containers as small as 11\,mm in diameter.}
      \label{fig_concept}
   \end{figure}

Several studies have attempted to address these challenges, but existing approaches typically solve either the scooping problem or the weighing problem, not both simultaneously.  
Previously, we proposed SCU-Hand~\cite{takahashi2025scuhand}, demonstrating that a soft conical end effector could exploit its inherent flexibility and variable size to scoop powders from containers of different shapes and sizes while compensating for positional errors using only position control. However, this method did not incorporate weighing and could only release the entire scooped amount.  
In terms of weighing, Jiang et al.~\cite{jiang2023autonomousbiomimetic} developed a dual-arm robot equipped with three types of spatulas to achieve weighing accuracy as low as 2\,mg. Their approach enabled scooping and weighing in the range of 20--1000\,mg, but the scooping was limited to large containers with abundant powders, and frequent tool switching made the transfer of larger quantities inefficient.  
These examples highlight that existing methods cannot simultaneously achieve flexibility, efficiency, and precise weighing functionality. To address this gap, this study proposes a novel design that preserves the advantages of softness and simplicity of the previous SCU-Hand~\cite{takahashi2025scuhand} while adding a weighing function. 

The key insight of this work is that the conical shape of the previous hand design~\cite{takahashi2025scuhand} resembles a funnel commonly used to pour powders into narrow-mouthed containers. Many containers employed in synthesis workflows are relatively small, with openings of around a few centimeters, whereas the SCU-Hand has a larger size of approximately 100\,mm. If powders can be dispensed from the apex of the cone, it becomes straightforward to deliver them into such small containers, as illustrated in Fig.~\ref {fig_concept}.  
Furthermore, to cope with the nonlinear nature of powder flow, we introduce an algorithm that incorporates a model commonly used for hopper flow prediction~\cite{wikacek2023converging}, and through online parameter identification, the system adapts to variations in powder dynamics.

The contributions of this work are summarized as follows:  
\begin{enumerate}
    \item We propose SCU-Hand-SV (\textbf{S}oft \textbf{C}onical \textbf{U}niversal \textbf{Hand} with integrated \textbf{S}ingle-sheet \textbf{V}alve) for robotic powder weighing. 
    \item We establish and validate a model that captures the nonlinear tendency of powder flow.
    \item We introduce a model-based control with online parameter identification that significantly improves accuracy and convergence compared to direct PID control.
\end{enumerate}

\section{Preliminary}

\subsection{Target Task}
As an initial investigation, this study focuses on transferring ground powders with relatively free-flowing and non-cohesive properties from a mortar, which is often used for grinding small quantities of material, into containers for subsequent processes, since existing methods already automate the pre-processing step of grinding. Before grinding, raw materials are dispensed from individually stored precursors, a task that can already be achieved using existing automation products~\cite{christensen2021automation}. In addition, dedicated grinding machines are often employed for this step~\cite{rajput2015methods}.  

Compared to dispensing before grinding, transferring ground material is characterized by a broader weighing range and relatively modest precision requirements. In the following we define the target weighing range and tolerance for the proposed system by presenting representative tasks, noting that these values depend on the specific method.



\subsection{Representative Tasks and Required Amounts}


\textbf{Measurement}: \textbf{tens to hundreds of milligrams}. 
Various analytical methods are used to measure material properties~\cite{patel2025matering,wei2014review,alsac2025characterizing}. Tens to hundreds of milligrams are often sufficient. A prime example is XRD analysis, one of the most frequently used measurement techniques~\cite{ali2022xray}, where powders are dispensed into a glass holder. Using smaller quantities is beneficial for high-throughput analysis. Reliable XRD measurements have been reported from as little as 3--80\,mg, with standard deviations below 1\%~\cite{Yotsumoto2024}. To ensure robust and stable reproducibility, this study adopts a lower bound of 20\,mg based on the findings in~\cite{Yotsumoto2024}.

\textbf{Synthesis}: \textbf{hundreds of milligrams to gram scale}. 
In materials synthesis, thermal processing methods often consist of two main steps: calcination and sintering~\cite{west2014solid}. Calcination involves heating a material to remove unwanted substances and is often performed on a sub-gram to gram scale.
Sintering then aims to create a dense, chemically bonded solid from the powder, usually preceded by a pelletizing step to increase density, often in the range of several hundred milligrams.
    
    
    
    While the details and sequence of these steps vary by study, some examples illustrate the process. For instance, in a study on solid electrolytes~\cite{kovsir2022comparative}, a 5\,g precursor powder was prepared in the first step, followed by the second step where a 13\,mm pellet was formed by uniaxially pressing 0.5\,g of the powder. Another example is a solid-state synthesis using platinum crucibles~\cite{zhang2016crystal}, where a single batch of multi-gram material was processed through both first and second steps.

\subsection{Target Specifications}

Although certain tasks, such as material measurements and synthesis, do not intrinsically require sub-milligram precision, a unified system should support a broad range of operations. For example, Yotsumoto et al.~\cite{Yotsumoto2024} showed that standard deviations below 1\% were achieved across 3--80\,mg samples, indicating that extremely fine precision is not always necessary. Nevertheless, to ensure adaptability to diverse workflows, we set the target tolerance in this study to be consistent with prior robotic weighing research~\cite{jiang2023autonomousbiomimetic}, namely $\pm$2\,mg.  

Based on these considerations, the weighing functionality targeted in this work is defined as handling \textbf{20\,mg to several grams, with an error less than $\pm$2\,mg}, thereby covering a wide range of measurement methods and material synthesis tasks within a single system.

\section{Related Work}\label{sec_related_work}
In this section, we review prior studies to determine whether any existing technologies can address the challenges targeted in this work. We then compare single-component flexible morphing mechanisms that could be integrated into the previous SCU-Hand~\cite{takahashi2025scuhand}.

\subsection{Approaches for Robotic Powder Weighing}
Various approaches have been proposed for robotic powder weighing, ranging from dedicated commercial products to general-purpose robotic manipulation methods.  

Representative commercial systems include Quantos (Mettler-Toledo International Inc.) and GDU-S SWIL (Chemspeed Technologies AG). Quantos employs screw and valve mechanisms to transport powders incrementally from preloaded dispensing heads~\cite{christensen2021automation}. Quantos uses a spiral screw to push powders, enabling dispensing across a wide range from milligrams up to approximately 100\,g. The screw mechanism also crushes agglomerates during feeding, thereby improving dispensing stability.  By contrast, GDU-S SWIL uses a glass capillary to aspirate and release powders of less than 30\,mg. While these systems achieve high precision, they are limited to handling very small quantities.
Other approaches include vibration-based powder feeder~\cite{3PInnovation_VibratoryFeeder}. Since powders are transported without direct contact, the surfaces that interact with powders are simple, and the system can dispense powders with poor flowability that do not flow under their own weight. However, their weighing precision is generally lower than that of screw-based systems. 

Research inspired by human scooping operations has also been reported. Schenck et al.~\cite{schenck2017learning} demonstrated automated powder scooping and dumping using a learning-based approach. Kadokawa et al.~\cite{kadokawa2023learning} achieved 0.1\,mg weighing precision in the 5--15\,mg range using a robot equipped with a very small spatula. Clarke et al.~\cite{clarke2018learning} attached a microphone to a custom scoop and estimated powder mass from acoustic signals, enabling weighing of relatively coarse granular materials such as plastic pellets and coffee beans.  

Although these spatula-based methods achieve high accuracy with simple tools, they are restricted to large containers filled with abundant powders. The amount that can be scooped at one time and the achievable weighing precision are constrained by spatula size, requiring multiple tools of different sizes to cover a wide weighing range~\cite{jiang2023autonomousbiomimetic}.  

\subsection{Soft Morphing Mechanisms}
Soft robotic morphing mechanisms that could be potentially applied to powder dispensing have also been investigated. These mechanisms typically aim to realize the opening and closing of gaps using thin, flexible structures.  

As for special actuators, dielectric elastomers (DEA)~\cite{xu2021deasoftvalve} and piezoelectric~\cite{li2021piezoreview}\cite{durasiewicz2021piezoelectric} devices are capable of deforming thin sheets, enabling large displacements while maintaining softness and thinness. However, they often require high voltages and may increase implementation costs.  

As a structural approach, Keely proposed a kirigami-inspired deformable spatula~\cite{keely2024kiri}, in which slits cut into a sheet expand under tension. The same principle could potentially be applied to valve mechanisms, although powder may leak through the gaps even in the closed state. 
A soft valve based on the Kresling origami structure has also been introduced~\cite{lu2022conicalKreslingorigami}, where a tubular origami twists to close its inner hole, but this design suffers from increased thickness and difficulty in miniaturization.  
Other structural approaches rely on the relative motion of multiple sheets. Yamada et al.~\cite{yamada2017loop} proposed a method in which two plates are displaced linearly to generate a loop-like deformation. Another approach uses plates connected by a rotational joint~\cite{xiong2023rapid}, where rotational input induces deformation of the overall structure.  

\subsection{Summary}
In summary, commercial systems and spatula-based robotic methods are limited by container geometry and powder availability. To date, no approach has successfully combined efficient scooping from diverse containers with stable weighing across a wide dispensing range. 
To preserve the advantages of previous work that can be constructed from a single flexible sheet, our study adopts a structural approach based on linear displacement to create loop-shaped deformations~\cite{yamada2017loop}, enabling controllable powder dispensing from the apex of a conical hand, while other flexible mechanisms face challenges of cost, leakage, or size. The details of the design are described in Sec.~\ref{sec_compalison_of_design}.

\section{SCU-Hand with Single-Sheet Valve}


In this section, we describe the design concept of the proposed funnel-shaped hand. To realize a system that enables both efficient scooping and accurate weighing, the following five functional requirements are identified:

\begin{enumerate}
  \item \textbf{Reconfigurable size}: The end effector must change its size according to the target container.
  \item \textbf{Soft adaptability}: The end effector must retain sufficient softness to adaptively deform to different container geometries and positional variations.
  \item \textbf{Valve mechanism at the cone apex}: The hand must incorporate a valve mechanism that enables stepwise dispensing of powder exclusively from the apex of the cone, allowing transfer into containers with a small diameter.
  \item \textbf{Adaptability to diverse powders}: Since powders exhibit various physical properties and flow dynamics, the hand must accommodate multiple dispense modes (e.g., gravity flow and vibration-assisted dispense).
  \item \textbf{Weighing function}: The system must measure the dispensed powder and control the dispensed weight by referencing the error from the target weight.
\end{enumerate}

Requirements (1) and (2) are inherited from the previous SCU-Hand implementation. In this chapter, we mainly discuss the design of the valve mechanism (requirement 3). The valve must be implemented without compromising requirements (1) and (2). Furthermore, in the implementation section, systems corresponding to requirements (4) and (5) will be described.

\subsection{Flexible Valve Mechanism}

\subsubsection{Requirements for the Valve Mechanism}

The following conditions are defined as necessary for the valve mechanism:

 \textbf{Simplicity}: The powder-contacting part is preferably composed of a single component. In material synthesis, contamination prevention is critical, which favors disposable or easy-to-clean designs. In this paper, we define ``simplicity'' as minimizing powder-contacting components and enabling low-cost fabrication. Our design adopts a single flexible sheet.

\textbf{Dispense from the cone apex}: For applications requiring powder to be transferred into containers with narrow openings, it is crucial that powder can be dispensed precisely from the apex of the cone. This ensures accurate delivery into confined geometries.  

\textbf{Variable opening size}: According to the powder flow model~\cite{wikacek2023converging}, the mass flow rate is determined by the size of the outlet. Therefore, for efficient weighing over a wide range of target quantities, the outlet must be adjustable to expand its cross-sectional area as required.  

   \begin{figure}[t]
      \centering
     \includegraphics[width=0.85\linewidth]{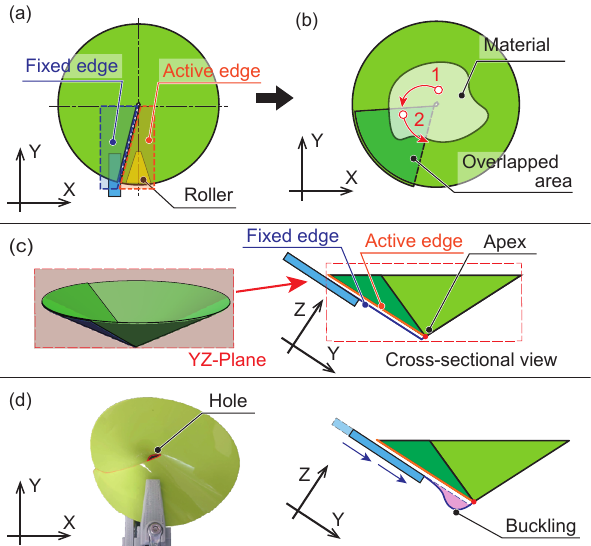}
      \caption{Proposed flexible valve: (a) Top-view of conical structure, (b) powder pass through between two sheets, (c) cross-sectional view of conical structure and two overlapping sheets, (d) buckling deformation}
      \label{fig_deformation_comparison}
   \end{figure}
   
\subsubsection{Concept of Valve Designs}\label{sec_compalison_of_design}

To satisfy the requirements described above, this study adopts the deformation method illustrated in Fig.~\ref{fig_deformation_comparison}. The main features of the proposed mechanism are as follows.  

\textbf{Simple design with single-sheet structure}: 
The sheet in contact with powders is designed to be disposable, enabling mass production by cutting a single sheet into shape. Unlike many conventional valve mechanisms, which generate gaps through the separation of two parts, the proposed mechanism employs a single-sheet conical structure overlapped with two sheets, a fixed edge on the left side of the slit, and an active edge driven by the rollers (Fig.~\ref{fig_deformation_comparison}~(a)). 
This overlap is treated as the valve, and by moving one of the sheet, a gap is formed that allows powders to be dispensed. The principle of this operation is illustrated in Fig.~\ref{fig_deformation_comparison}~(b). When a gap is formed, powder first enters the inter-sheet space and then exits through the gap, thereby enabling dispensing.

\textbf{Independent from reconfigurable mechanism}: 
To avoid situations in which the valve cannot operate at specific cone sizes, the proposed design moves the fixed edge rather than the active edge. As a result, powder dispensing can be performed independently of the reconfiguration motion. As shown in the cross-sectional view in Fig.~\ref{fig_deformation_comparison}~(c), the fixed edge lies below the active edge, and by displacing the fixed edge in the Z direction, a gap is created between the two sheets.  

\textbf{Buckling-like hole formation near the apex}:   
By exploiting the buckling deformation of the sheet (Fig.~\ref{fig_deformation_comparison}~(c)), a gap can be generated near the cone apex. This deformation is realized by sliding the fixed edge along the Y direction, and the sliding displacement determines the size of the gap. Consequently, the mechanism can produce large deformations and enables precise powder transfer even into small containers. 
The procedure for reproducing this deformation is described in the Appendix and is helpful for understanding the principle of the flexible valve.

\subsection{Implementation of Powder Weighing System}

\begin{figure}[t]
      \centering
     \includegraphics[width=0.9\linewidth]{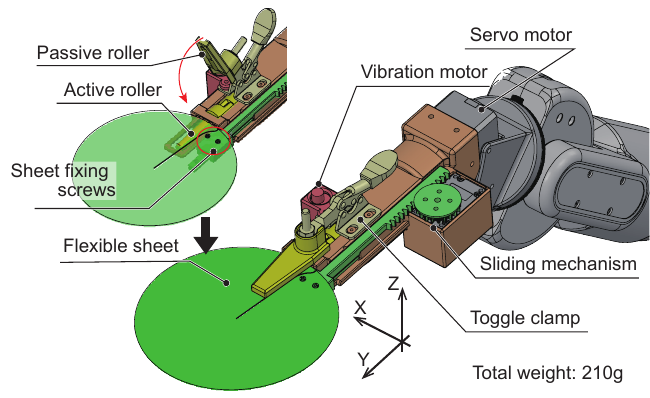}
      \caption{Design of the proposed SCU-Hand-SV. The flexible sheet can be easily replaced by removing two screws.}
      \label{fig_hand_design}
   \end{figure}

This section describes the detailed mechanical structure of the proposed end effector and the weighing control system. The overall design is shown in Fig.~\ref{fig_hand_design}. The weight of the end efector is 210\,g. While the mechanism remains close to the previous conical sheet structure~\cite{takahashi2025scuhand}, several modifications have been introduced to incorporate weighing capability.  

  \begin{figure}[t]
      \centering
     \includegraphics[width=0.7\linewidth]{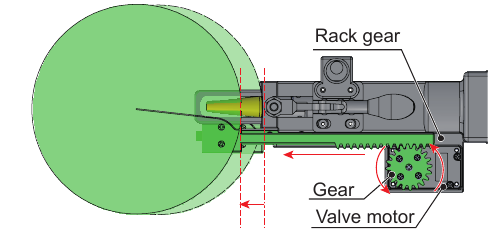}
      \caption{Sliding mechanism of the flexible valve for radial displacement.}
      \label{fig_slide_mechanism}
   \end{figure}

\subsubsection{Mechanical Design}
The proposed system consists of three main elements: the \textit{Morphing mechanism}, the \textit{Flexible valve}, and the \textit{Vibration motor}, all mounted at the tip of a robotic arm (CRANE-X7, RT Corporation). The coordinate system is defined with the flexible sheet surface as the XY plane, and the perpendicular direction as the Z. 

\textbf{Morphing mechanism}: 
As in the previous work~\cite{takahashi2025scuhand}, the flexible sheet is clamped between two rollers and driven by friction. The rollers actuate the active edge on the right side of the slit of the sheet. One roller is connected to a servo motor (Dynamixel XM430-W350-R, ROBOTIS Co., Ltd.) through a shaft to transmit rotational motion. The flexible sheet is a 0.2\,mm white polyacetal sheet cut into a circular shape with a diameter of 100\,mm. However, in the figures of this paper, a green polypropylene sheet was used for better visibility.
In the design of this paper, the other passive roller is held by a toggle clamp, allowing it to be easily detached from the sheet. With this structure, the sheet can be replaced by removing only two screws, which enables replacing the flexible sheet quickly.  

\textbf{Flexible valve}: 
The major design modification introduced in this work is the flexible valve. To realize the radial deformation mechanism described in~\ref{sec_compalison_of_design}, the fixed edge is driven along the Y direction (Fig.~\ref{fig_slide_mechanism}). The fixed edge is attached to a rod-like component supported by a linear guide. The opposite side of this component is shaped as a rack gear, which meshes with a gear connected to a valve motor(Dynamixel XC330-T288-T, ROBOTIS Co.,Ltd.) mounted on the side of the end effector, enabling forward and backward motion. Through the combination of the frictional fixation of the active edge and the reciprocating movement of the fixed edge, the sheet deforms in the Z direction, thereby allowing powder to be dispensed from the apex of the cone (Fig.~\ref{fig_concept}~(c)).  

\textbf{Vibration motor}:   
Certain powders cannot be dispensed solely by gravity, as they may clog near the outlet. To address this, a vibration-assisted dispensing function is included. The actuator is a small DC motor (N20 motor) with an eccentric weight attached directly to its rotating shaft. The motor shaft is aligned parallel to the Z axis, generating vibrations that improve powder flowability and stabilize dispensing even for cohesive powders. We generally do not use the vibration function, except when dispensing small amounts of powder that do not flow well under gravity. 

\subsubsection{Weighing System}

The proposed method integrates the end effector with an electronic balance. The balance is connected to a PC via serial communication and continuously transmits measurement data.
The PC receives the data and determines the control strategy for the manipulator.  
The valve motor and vibration motor of the end effector are controlled by a microcontroller that is also connected to the PC via serial communication. The PC sends commands including the valve motor rotation angle, valve opening duration, and vibration activation, which are interpreted by the microcontroller to actuate the end effector. Through this integration, dispensed quantities are monitored in real-time, and feedback control is applied to converge on the target mass.  


   \begin{figure}[t]
      \centering
     \includegraphics[width=0.8\linewidth]{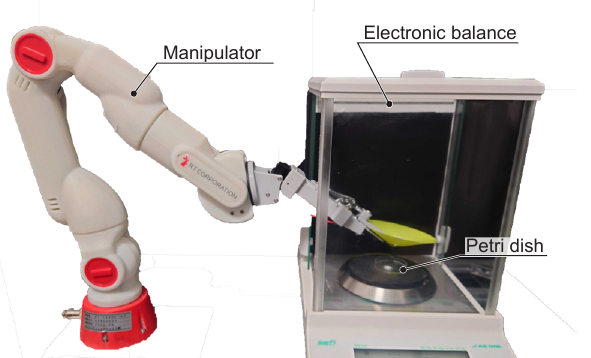}
      \caption{Experimental setup: the proposed hand positioned in the electronic balance, dispensing powders into a petri dish placed on the balance.}
      \label{fig_experimental_setup}
   \end{figure}

\subsubsection{Powder Weighing Control}\label{sec_weighing_control}

To handle powders with diverse physical properties, the proposed system employs a weighing control algorithm that combines model-based dispensing prediction with online parameter identification (Algorithm~\ref{alg_weighing_control}).  

\textbf{Model formulation}: 
The model is derived from the empirical Beverloo equation, which describes the mass flow rate $Q$ of powders through a hopper outlet~\cite{wikacek2023converging}:  
\begin{equation}
    Q = C \, \rho_b \, \sqrt{g} \, \left( D_o - k \, d \right)^{\frac{5}{2}}\label{eq_beverloo}
\end{equation}

where $C$ is an empirical coefficient, $\rho_b$ is the bulk density, $g$ is the gravitational acceleration, $D_o$ is the outlet diameter, $d$ is the particle diameter, and $k$ is the particle correction factor.  

\textbf{Simplification and dispensing model}:   
In our implementation, the particle correction term $k d$ is neglected by assuming a sufficiently small particle size, and powder-specific constants are grouped into a single coefficient $C'$. The dispensed mass $W_{\text{drop}}$ is then expressed as  
\begin{equation}
    W_{\text{drop}} = Q \, t = C' \, D_o^{2.5} \, \bigl( T(D_o) + t_{\text{pose}} \bigr),\label{eq_target_drop}
\end{equation}

where $C'$ is a lumped correction coefficient capturing powder-specific properties, $T(D_o)$ is the time required for the valve to reach the target opening size, and $t_{\text{pose}}$ is the waiting time after the valve reaches the desired opening. This formulation provides a practical approximation for predicting dispensing amounts across a wide range of outlet sizes and timings.  

\textbf{Parameter identification and control}:   
During operation, the commanded outlet diameter $D_o$, the waiting time $t_{\text{pose}}$, and the measured dispensed mass $W_{\text{drop}}$ are recorded. Using these values, the coefficient $C'$ is iteratively updated via least squares estimation, enabling the model to adapt online to variations in powder properties and environmental conditions.  

After model fitting, control commands are generated according to Algorithm~\ref{alg_weighing_control}. These include valve opening size, duration, and vibration activation. By applying the model-based prediction with online updates, the system achieves real-time correction of deviations from the target mass, ensuring stable and accurate weighing performance.
The vibration function is employed only when the estimated powder drop does not reach the target weight even with the maximum value of valve size and waiting time.

\begin{algorithm}
\caption{Dispensing Control (abstract, implicit vibration switch)}
\label{alg_weighing_control}
\begin{algorithmic}
  \Require $W_{\text{measured}},\, W_{\text{goal}},\, K_p$
  \State $W_{\text{error}} \gets W_{\text{goal}} - W_{\text{measured}}$
  \State $use\_vibration \gets \textbf{false}$
  \While{$|W_{\text{error}}| \ge 2$}
    \If{$\Delta W \ge 0.5$}
      \State \textbf{Store} $(L_{deff},\, t_{pose},\, \Delta W)$
      \State \textbf{Fit} parameter $C'$ by least squares on stored pairs
      \Comment{$\;\Delta W \approx W\_drop(L, t;\, C')$}
    \EndIf

    \State $W_{\text{target}} \gets K_p \cdot W_{\text{error}}$
    \State $(L^\star, t^\star) \gets \arg\min_{L,t}\; \big|\, W\_drop(L, t;\, C') - W_{\text{target}} \,\big|$

    \State $W_{\text{test}} \gets W\_drop(210,\,200;\, C')$
    \If{$W_{\text{test}} < W_{\text{target}}$}
      \State $use\_vibration \gets \textbf{true}$ 
      \Comment{model $W\_drop(\cdot;\, C')$ is switched internally to vibration mode}
    \EndIf

    \State \textbf{Send to Arduino}: $Valve\_angle \gets L^\star,\; t\_pose \gets t^\star$; \textbf{Execute} Dispense
    \State \textbf{Measure} new $W_{\text{measured}}$; \; $W_{\text{error}} \gets W_{\text{goal}} - W_{\text{measured}}$
  \EndWhile
\end{algorithmic}
\end{algorithm}

\section{Experimental Validation}

We conducted a comprehensive experimental evaluation to assess three critical aspects of the proposed system: dispensing precision, effectiveness across different powder materials, and flow rate predictability. Our experiments included performance evaluations using various powder types and comparative analysis with conventional weighing control methods.

\subsection{Experimental Setup}
\begin{figure}[t]
      \centering
     \includegraphics[width=0.8\linewidth]{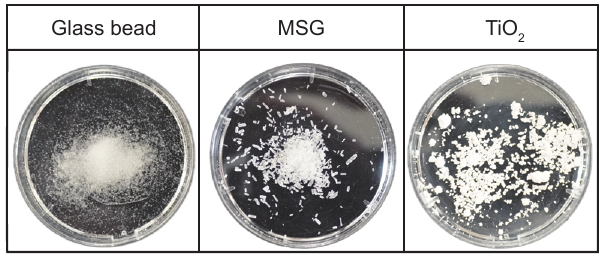}
      \caption{Material used in experiment}
      \label{fig_material}
   \end{figure}

All experiments were conducted using the weighing system described in Section~\ref{sec_weighing_control}. The flexible conical sheet had an initial diameter of 100\,mm, and a portion of the sheet was overlapped so that the diameter of the bottom circle of the cone became 90\,mm. As shown in Fig.~\ref{fig_experimental_setup}, the axis of the cone was aligned vertically, and the tip of the conical structure was positioned 80\,mm above the measurement surface of the electronic balance by adjusting the joints of the CRANE-X7 robotic arm. As powder materials, we used three types shown in Fig~\ref{fig_material}: glass beads (AS ONE CORPORATION, with a diameter of approximately 0.1\,mm), monosodium glutamate (MSG; Ajinomoto Co., Inc.), and titanium dioxide (TiO$_2$; Kojundo Chemical Lab. Co., Ltd.). The amount of powder loaded into the conical structure was 5000\,mg in most cases, but for TiO$_2$, which has a relatively low bulk density, the quantity was limited to 4000\,mg to fit. Dispensed powders were delivered into a Petri dish. 
The target weights were set to 20\,mg, 50\,mg, 500\,mg, and 3000\,mg.  

At the start of each trial, a fixed quantity of powder was placed in the conical structure, after which the system dispensed powder until the desired target amount was achieved. Following each trial, the dispensed powder was returned to the structure, and the next trial was performed. For each condition, 10 trials were conducted. The control algorithm was identical across experiments, with only the proportional gain and maximum valve motor rotation adjusted for each powder. As a baseline, we implemented a direct PID control method for only dispensing glass beads, where the valve motor angle and opening time were determined by a simple PID loop based on the error between the current and target weights. 
The direct PID parameters were primarily tuned for the 500\,mg case, and the same parameters were subsequently applied to the minimum case of 20\,mg and the maximum case of 3000\,mg; the 50\,mg case was not tested. The maximum number of control steps was limited to 100, and if convergence was not achieved within this limit, the control sequence was terminated.

\subsection{Results}
The results of Experiment~1 are summarized in Tables~\ref{tab_sucess_rate} and~\ref{tab_time_and_step}. Table~\ref{tab_sucess_rate} shows the number of successful trials within $\pm$2\,mg, together with the mean and standard deviation of the final dropped weights. Table~\ref{tab_time_and_step} shows the average and standard deviation of the number of control steps and the total time required for each trial. In the visualization, trials in which all ten attempts were successful are highlighted in red, those with success rates above 80\% are highlighted in yellow, and those below 80\% are highlighted in blue.
Direct PID indicates the result of glass beads with direct PID control.

Glass beads and MSG consistently achieved success rates above 80\% across all target amounts. TiO$_2$, however, showed success rates below 50\% for targets between 50 and 3000\,mg, although in most failed cases the error remained within about 20\,mg. With the proposed algorithm, 80\% of all trials achieved an error within $\pm$2\,mg, with the maximum error observed being approximately 20\,mg for TiO$_2$. Direct PID control succeeded only for the 500\,mg case, achieving 100\% successful trials. At 20\,mg, six of the ten trials did not converge within the maximum step limit, while at 3000\,mg, all trials overshot the target.  

\begin{figure}[t]
      \centering
     \includegraphics[width=0.75\linewidth]{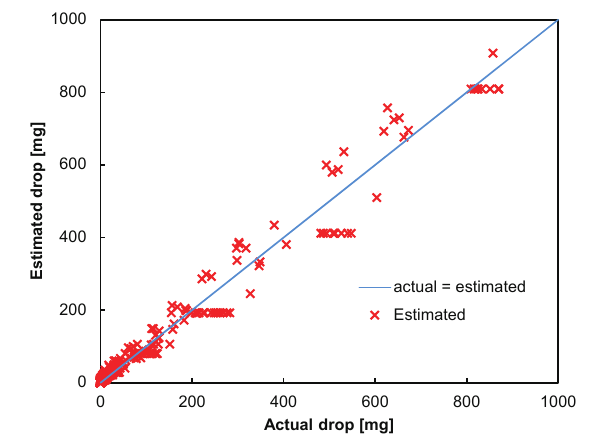}
      \caption{Comparison between model-based estimated value and actual drop}
      \label{fig_fitting_graph}
   \end{figure}

In terms of weighing time, the proposed algorithm consistently completed trials within a few minutes for all target amounts. In contrast, direct PID control required significantly longer times, depending on the condition. The use of vibration was automatically determined by the algorithm according to the powder characteristics: for glass beads, vibration was never applied; for MSG, vibration was applied only in some steps; and for TiO$_2$, vibration was applied in all steps because dispensing could not proceed at all without it.  
Fig.~\ref{fig_fitting_graph} illustrates the relationship between the model predictions and the experimental measurements. In the plot, the horizontal axis represents the measured values, and the vertical axis represents the dispense amounts obtained from Eq.~\ref{eq_target_drop} using the corresponding control parameters. If the plotted points lie close to the line of equality between measured and predicted values, the model can be considered accurate. All data used in this plot were obtained from the experiments with glass beads, and the parameter $C'$ was fitted to the entire dataset using the least-squares method. The resulting coefficient of determination was $R^2 = 0.98$, indicating that the model provides a highly reliable prediction of powder dispensing behavior.

Our experimental results demonstrate the system's effectiveness across three critical aspects: First, the proposed conical structure achieved a weighing error ranging from about $\pm$2\,mg to 20\,mg even with its simple mechanics, confirming that the target accuracy is attainable. 
Second, the system effectively handled different powder materials by automatically determining when vibration assistance was necessary. Third, the predictive flow model significantly improved both accuracy and convergence speed compared to conventional weighing control methods, addressing the highly nonlinear relationship between control parameters and dispensed amounts.

\setlength{\tabcolsep}{3pt}
\begin{table}[t]
  \centering
  \caption{Success Rate and Dropped Weight in the Experiment}
  \label{tab_sucess_rate}
  \begin{tabular}{lcccc}
    \toprule
    & 20\,mg & 50\,mg & 500\,mg & 3000\,mg \\
    \midrule
    \textbf{Glass beads} & & & & \\
    ~Success rate & \cellcolor[HTML]{FFFC9E}9/10  & \cellcolor[HTML]{FFFC9E}9/10  & \cellcolor[HTML]{FFCCC9}10/10 & \cellcolor[HTML]{FFCCC9}10/10 \\
    ~Dropped [mg] & $20.0 \pm 1.3$               & $49.9 \pm 2.5$               & $498.6 \pm 0.4$              & $2998.8 \pm 0.67$                          \\
    \midrule
    \textbf{MSG} & & & & \\
     ~Success rate & \cellcolor[HTML]{FFCCC9}10/10 & \cellcolor[HTML]{FFCCC9}10/10 & \cellcolor[HTML]{FFFC9E}8/10  & \cellcolor[HTML]{FFCCC9}10/10 \\
     ~Dropped [mg] & $18.6 \pm 0.5$               & $43.9 \pm 0.5$               & $500.5 \pm 1.9$              & $2999.1 \pm 1.5$                          \\
    \midrule
    \textbf{TiO$_2$} & & & & \\
     ~Success rate & \cellcolor[HTML]{FFFC9E}8/10  & \cellcolor[HTML]{96FFFB}5/10  & \cellcolor[HTML]{96FFFB}4/10  & \cellcolor[HTML]{96FFFB}4/10  \\
     ~Dropped [mg] & $20.2 \pm 2.8$               & $52.4 \pm 3.5$               & $503.5 \pm 3.5$              & $3003.6 \pm 2.8$                          \\
    \midrule
    \textbf{Direct PID} & & & & \\
     ~Success rate & \cellcolor[HTML]{96FFFB}4/10  & N/A                           & \cellcolor[HTML]{FFCCC9}10/10 & \cellcolor[HTML]{96FFFB}0/10  \\
     ~Dropped [mg] & $16.2 \pm 2.3$               & N/A                           & $498.9 \pm 0.8$              & $3423.0 \pm 244.6$                        \\
    \bottomrule
  \end{tabular}
\end{table}
\setlength{\tabcolsep}{6pt}

\begin{table}[t]
  \centering
  \caption{Step counts and execution time at different target masses.}
  \label{tab_time_and_step}
  \begin{tabular}{lcccc}
    \toprule
    & 20\,mg & 50\,mg & 500\,mg & 3000\,mg \\
    \midrule
    \textbf{Glass beads} & & & & \\
    ~Step     & $8 \pm 0.9$   & $9 \pm 1.7$   & $19 \pm 3.3$  & $19 \pm 1.1$ \\
    ~Time [s] & $59 \pm 9$    & $65 \pm 15$   & $141 \pm 56$  & $173 \pm 11$ \\
    \midrule
    \textbf{MSG} & & & & \\
    ~Step     & $14 \pm 3.7$  & $17 \pm 3.4$  & $24 \pm 1.9$  & $36 \pm 4.0$ \\
    ~Time [s] & $49 \pm 33$   & $158 \pm 28$  & $205 \pm 19$  & $327 \pm 38$ \\
    \midrule
    \textbf{TiO$_2$} & & & & \\
    ~Step     & $13 \pm 0.8$  & $14 \pm 0.5$  & $19 \pm 0.6$  & $39 \pm 5.6$ \\
    ~Time [s] & $106 \pm 10$  & $123 \pm 10$  & $164 \pm 6$   & $344 \pm 64$ \\
    \midrule
    \textbf{Direct PID} & & & & \\
    ~Step     & $90 \pm 28.7$ & N/A           & $38 \pm 6.7$  & $11 \pm 0.5$ \\
    ~Time [s] & $683 \pm 216$ & N/A           & $369 \pm 76$  & $103 \pm 6$ \\
    \bottomrule
  \end{tabular}
\end{table}

\section{Discussion}

The experiments demonstrated that the proposed flexible valve mechanism combined with the weighing control algorithm can achieve effective precision at the milligram level. 
For glass beads and MSG, most cases fell within an error of $\pm$2\,mg, while TiO$_2$ showed lower precision, with a maximum error of 20\,mg.
These findings confirm that the proposed conical structure is fundamentally effective for diverse powder materials.  
The present system demonstrated a weighing range of 20\,mg to 3\,g. Scaling to larger amounts would require increasing the quantity of powder loaded into the conical structure, but as observed in the case of TiO$_2$, powders with low bulk density risk spilling over. The effects of using larger sheets on weighing accuracy and dispensing stability remain important subjects for future investigation.  

Two main directions for improvement emerged. The first is improving control accuracy. TiO$_2$, which could not be dispensed by gravity alone, always required vibration to be applied. While vibration-assisted dispensing is effective for cohesive powders, it also introduces variability. Future work should consider including vibration intensity and frequency as additional control parameters to reduce variation and further enhance accuracy.  

The second is reducing control time. In the current algorithm, each control step requires, on average, about 10 seconds. The main cause of this delay was the waiting time until the balance readings stabilized after powder was dispensed. This waiting period is largely dependent on the hardware performance of the electronic balance. With faster balances or advanced filtering of measurement data, the proposed system has the potential to complete the weighing process in a much shorter time.

\section{Conclusion}

This paper presented a novel end effector, SCU-Hand-SV,  and a weighing system to improve powder handling for the full automation of small-scale solid-state synthesis workflows. The proposed end effector builds on the previously reported conical flexible hand, to which we added an outlet at the apex of the cone. This design enables incremental dispensing while preserving the softness and simplicity of the original flexible sheet. By integrating the end effector with an external electronic balance in a feedback system, we achieved milligram-level weighing precision across powders with different flow properties.  
A major advantage of the proposed mechanism is its adaptability to diverse tasks. The proposed method can potentially be extended beyond powders to liquids, viscous fluids, or slurries, though accuracy may need to be relaxed in cases where surface tension dominates (see supplementary video). Furthermore, handling cohesive or sticky materials remains an important direction for future work.

In summary, the proposed system provides a novel approach to robotic powder dispensing, demonstrates milligram-level precision with different powders, and shows promise for adaptation to broader tasks and materials. It has the potential to bring innovative advances to automated processes in diverse fields.

\addtolength{\textheight}{-12cm}   



\section*{APPENDIX}
\begin{figure}[h]
      \centering
     \includegraphics[width=0.75\linewidth]{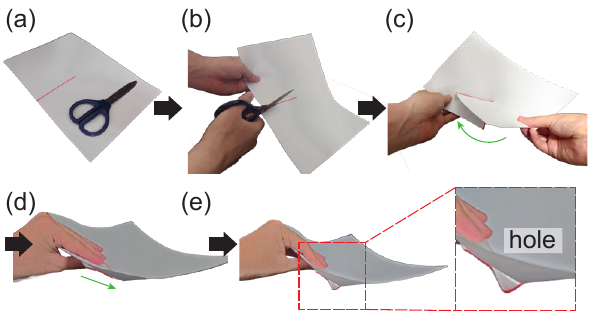}
      \caption{Reproducing the proposed mechanism by using paper: (a) Prepare a sheet of paper and scissors, (b) cut the paper halfway toword the center of paper, (c) overlap the right side of the slit onto the left side, (d) pinch the two overlapping sheets with fingers, and (e) slide the lower sheet toward the center to create a hole around the apex.}
      \label{fig_paper_reproduction}
   \end{figure}

This section explains the reproduction of the proposed mechanism using paper. A conical structure is created by making slits in paper of any shape and overlapping the two edges. The detailed procedure for the lower part is shown in the caption of Fig.~\ref{fig_paper_reproduction}.


\bibliographystyle{IEEEtran}
\bibliography{reference}

\end{document}